\documentclass[10pt,conference]{IEEEtran}
\IEEEoverridecommandlockouts

\usepackage{cite}
\usepackage{amsmath,amssymb,amsfonts}
\usepackage{algorithmic}
\usepackage{graphicx}
\usepackage{textcomp}
\usepackage{xcolor}
\usepackage{algorithm}
\usepackage{multirow} 
\usepackage{booktabs,tabularx, colortbl} 

\makeatletter
\def\ps@IEEEtitlepagestyle{%
  \def\@oddfoot{\mycopyrightnotice}%
  \def\@oddhead{\hbox{}\@IEEEheaderstyle\leftmark\hfil\thepage}\relax
  \def\@evenhead{\@IEEEheaderstyle\thepage\hfil\leftmark\hbox{}}\relax
  \def\@evenfoot{}%
}
\def\mycopyrightnotice{%
  \begin{minipage}{\textwidth}
  \centering \scriptsize
  Copyright~\copyright~2025 IEEE. Personal use of this material is permitted. Permission from IEEE must be obtained for all other uses, in any current or future media, including reprinting/republishing this material for advertising or promotional purposes, creating new collective works, for resale or redistribution to servers or lists, or reuse of any copyrighted component of this work in other works.
  \end{minipage}
}
\makeatother

\def\BibTeX{{\rm B\kern-.05em{\sc i\kern-.025em b}\kern-.08em
    T\kern-.1667em\lower.7ex\hbox{E}\kern-.125emX}}
\begin{document}

\title{WeightedKV: Attention Scores Weighted Key-Value Cache Merging for Large Language Models}

\author{
Jian Yuan$^{1\dagger}$ \thanks{$\dagger$ Equal contribution.},
Ziwei He$^{1\dagger}$, 
Haoli Bai$^{2}$, 
Jingwen Leng$^{1}$,
Bo Jiang$^{1*}$ \thanks{$*$ Bo Jiang is the corresponding author.}
\\
$^{1}$Shanghai Jiao Tong University
\quad $^{2}$Huawei Noah's Ark Lab
\\
\{yuanjian, ziwei.he, leng-jw, bjiang\}@sjtu.edu.cn
}

    \maketitle

\begin{abstract}

Large Language Models (LLMs) use key-value (KV) cache to reduce redundant computation in autoregressive generation. However, the KV cache size increases linearly during generation, leading to excessive memory usage, especially for long texts. Most KV cache compression methods evict the unimportant KV pairs to maintain a fixed cache size, which leads to the permanent loss of tokens during generation. 
However, singular value decomposition shows that \textit{values} do not exhibit a strong low-rank property as \textit{keys} do, suggesting that information is distributed more evenly across \textit{values}, in contrast to its more redundant distribution within \textit{keys}.
Therefore, methods that evict both \textit{keys} and \textit{values} risk losing crucial information and compromise context integrity, ultimately degrading the output quality. To address this problem, we propose WeightedKV, a novel, training-free approach that discards the \textit{keys} of less important tokens, while merging their \textit{values} into neighboring tokens via a convex combination weighted by their average attention scores.
In this way, the retained \textit{keys} serve as anchors that guide the generation process, while the merged \textit{values} provide a rich contextual backdrop. We assess our method on four widely used language modeling datasets, demonstrating superior performance compared to all baseline methods, particularly with a lower budget ratio.

\end{abstract}

\begin{IEEEkeywords}
Large Language Models, KV Cache Compression, Efficient Inference
\end{IEEEkeywords}

\section{Introduction}
\label{sec:intro}

Recently, the decoder-only transformer has emerged as the leading architecture for large language models (LLMs)\cite{touvron2023llama,jiang2023mistral,achiam2023gpt}, showcasing exceptional effectiveness across various application areas\cite{zhang2024benchmarking,kamalloo2023evaluating,roziere2023code}. Nevertherless,  inference is costly due to the autoregressive nature of LLMs, which  generate new tokens by repeatedly using key-value (KV) pairs of previous ones. To minimize the redundant calculations, these generated KV pairs are stored in a cache, known as KV cache\cite{pope2022efficiently}, for later use, allowing the models to trade memory for computational power.  However, as sequence length increases, the KV cache presents significant scalability challenges, leading to memory and latency bottlenecks for LLMs. 

\begin{figure}
\centering
\includegraphics[height=4cm]{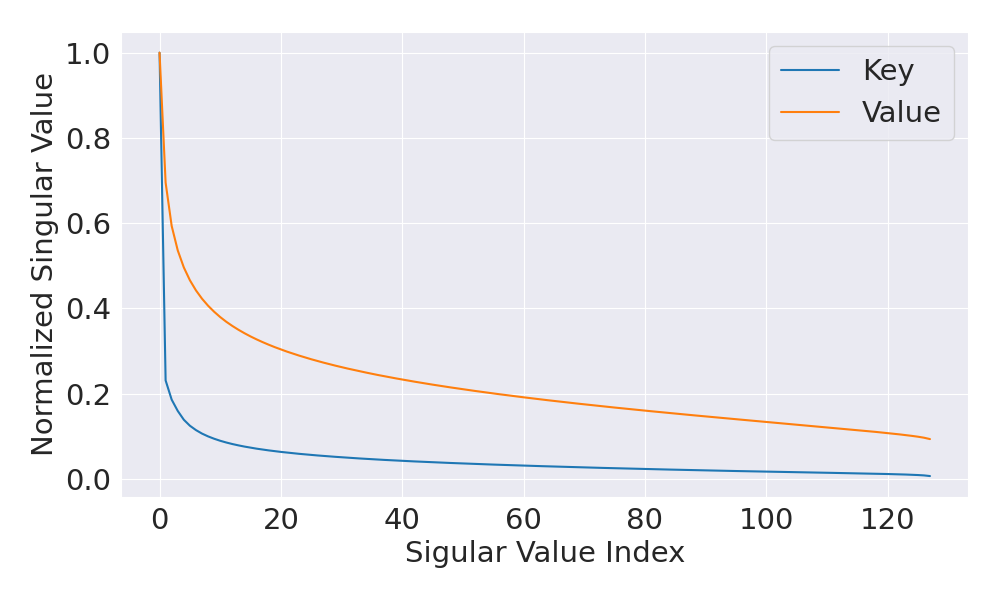}

\vspace{-2mm}
    \caption{Normalized singular values of KV averaged over the first 10 sequences truncated to length 1k in the PG19 test set.}
    \label{fig:singular}
\end{figure}

Several KV cache eviction methods have been proposed to optimize KV cache usage by maintaining a fixed size during generation. One approach involves static cache management, which evicts KV pairs of tokens outside predefined scopes, as seen in StreamingLLM\cite{xiao2023efficient} and LM-Infinite\cite{han2024lm}. Another approach involves dynamic management, where less important KV pairs are removed based on their impact on model performance\cite{zhang2024h2o, oren2024transformers, liu2024scissorhands}. While these methods ensure the KV cache remains within capacity, they inevitably lead to the permanent loss of entire KV pairs during generation. These eviction strategies could discard crucial KV pairs unintentionally, struggling to maintain context and diminish model's performance in long sequence generation.
To further investigate the effects of permanent eviction of both \textit{keys} and \textit{values}, we perform singular value decomposition on the hidden states of KV pairs across different layers and heads from Llama-2-7B\cite{touvron2023llama}.  Fig.\ref{fig:singular} shows the normalized singular values averaged over the first 10 sequences from PG19\cite{rae2019compressive}. 
Note that the normalized singular values for \textit{keys} rapidly approach zero, whereas those for \textit{values} exhibit a much heavier tail. This suggests that information within \textit{keys} is more redundant, while it is distributed more evenly across \textit{values}. 
Therefore, while it may be reasonable to discard the \textit{keys} of evicted tokens, simply discarding their \textit{values} may lead to information loss and degrade the quality of the generated outputs.

To address these issues, we propose WeightedKV, a novel, training-free approach that discards the \textit{keys} of less important tokens, while merging their \textit{values} into neighboring tokens via a convex combination weighted by their average attention scores. 
In contrast to eviction-based methods, WeightedKV does not simply evict KV pairs, but instead focuses on a more nuanced management of the cache. By maintaining a subset of \textit{keys} corresponding to important tokens, we ensure that critical information remains accessible without overloading device memory. The convex merging of the \textit{values} allows us to create a more compact representation that captures the diverse contributions of non-retained tokens, ensuring that the overall context is preserved even in their absence. In this way, the retained \textit{keys} serve as anchors that guide the generation process, while the merged \textit{values} provide a rich contextual backdrop. It is important to note that the convex combination of \textit{values} is derived from a theoretical analysis of how evicting \textit{keys} and merging \textit{values} can approximate an ideal merging that minimizes perturbations in the attention output. Additionally, we show that the proposed way of merging \textit{values} in a single step has only a minimal impact on the attention score distribution in future steps in Section \ref{sec: Methodology}.

Our experimental setup evaluates the performance of WeightedKV against some conventional eviction methods\cite{xiao2023efficient, zhang2024h2o, oren2024transformers} and another \textit{value} merging method CAM\cite{zhang2024cam} on four language modeling datasets. WeightedKV demonstrates the state-of-the-art performance in perplexity for long text generation compared to all baselines. By optimizing the KV cache with WeightedKV, we can not only reduce memory consumption but also improve the quality of generated sequences.

Our contributions can be summarized as follows:
\begin{itemize}
    \item We propose WeightedKV, an efficient, training-free approach that compresses the KV cache by preserving the \textit{keys} of important tokens, and merging the \textit{values} of discarded tokens into retained ones, via a convex combination weighted by their average attention scores. 
    \item We show that WeightedKV is grounded on a theoretical derivation of how evicting tokens and merging \textit{values }can approximate an ideal merging that minimizes perturbations in attention output. We also demonstrate that our \textit{value} merging incurs negligible perturbation to subsequent inference compared to full attention.
    \item We present thorough experimental results confirming the effectiveness of WeightedKV in the generation stage. It achieves the lowest perplexity on PG19, OpenWebText2, ProofPile, and ArXiv, compared to all baselines. 
\end{itemize}

\section{Related Work}
\label{sec:related}
There has been an increase in the number of works focusing on compressing KV cache during generation in the field of natural language processing, making it a critical area for enhancing the performance of LLMs with longer contexts. These methods can be categorized into eviction-based and merging-based approaches.

\paragraph{Eviction-based methods} 
To limit the size of the KV cache, LM-Infinite \cite{han2024lm} and StreamingLLM \cite{xiao2023efficient} utilize static cache management. They retain predefined initial and recent regions of a sequence, which typically receive high attention scores, while evicting KV pairs of tokens outside these regions. Other methods use more dynamic strategies that remove the least important KV pairs according to various importance measures. TOVA\cite{oren2024transformers} identifies the token with the lowest attention score from the previous step as the least important and prioritizes it for eviction. Recognizing that some tokens with high attention scores are crucial for model performance, H2O\cite{zhang2024h2o} determines token importance based on cumulative attention scores. Similarly, Scissorhands\cite{liu2024scissorhands} binarizes these cumulative scores to establish token importance.

\paragraph{Merging-based methods} 
KVMerger\cite{wang2024model} is a method that merges both the \textit{key} and \textit{value} states. Motivated by the observation that \textit{key} states show considerable token-level similarity, KVMerger first clusters the \textit{key} states into several groups using cosine similarity. Within each group, the \textit{key} and \textit{value} states are then merged according to their cosine similarity and attention scores, respectively. CaM \cite{zhang2024cam} shares some similarity with WeightedKV, as it also discards \textit{keys} and merges \textit{values}. However, CaM merges the \textit{values} of evicted tokens only probabilistically, with a non-negligible probability of discarding them and hence losing information.  
When a \textit{value} state is to be merged, it is scaled by a constant factor of $1/n$ 
and added to the \textit{value} states of the next $n$ tokens. In contrast, WeightedKV uses a deterministic merging process that accounts for the importance of different tokens as measured by their contributions to model performance.
There are also some other merging methods such as Dynamic Memory Compression (DMC)\cite{nawrot2024dynamic} and Anchor-LLM\cite{pang2024anchor}. Those methods  require additional parameter optimization and are not training-free.

\section{Preliminary}
\subsection{Background}
First, we illustrate the auto-regressive decoding process during generation. We use $\mathbf{W}_q, \mathbf{W}_k, \mathbf{W}_v\in\mathbb R^{d\times d}$ to denote the attention modules' weights in a layer. Here $d$ is the hidden dimension and we omit the dimensions of batches and heads for simplicity. Initial KV cache $\mathbf{K}_0, \mathbf{V}_0$ is set as empty.
Suppose at generation step $t$ a new input $\mathbf{x}_t \in \mathbb R^{d}$ enters, the attention gets \textit{query}, \textit{key}, and \textit{value} by
\[
\mathbf{q}_t= \mathbf{W}_Q \mathbf{x}_t,
\mathbf{k}_t= \mathbf{W}_K \mathbf{x}_t,
\mathbf{v}_t= \mathbf{W}_V \mathbf{x}_t.\]
Then the KV cache at the current step is updated by concatenating the new \textit{value} and \textit{key} to the cache 
\[
\mathbf{K}_t=\left[\mathbf{K}_{t-1},\mathbf{k}_t\right], 
\mathbf{V}_t=\left[\mathbf{V}_{t-1},\mathbf{v}_t\right].
\]

Using the updated KV cache, we can compute the attention weights $\mathbf{A}_t$ and the output of the attention module as
\[\mathbf{o}_t = \mathbf{A}_t^\top \mathbf{V}_t, 
\text{ where }
\mathbf{A}_t=\mathrm{softmax}\left(\mathbf{q}_t^{\top}\mathbf{K}_t\right).
\]
As the generation goes on, $\mathbf{K}_t,\mathbf{V}_t$ grows linearly with $t$. This allows the model to reference all previous tokens, but it comes with significant computing and storage costs as the context grows longer.

\subsection{Ideal Merging}
\label{sec: ideal merge}

We propose evicting \textit{keys} and merging \textit{values} based on the previous singular value analysis, demonstrating that this approach can achieve evicting \textit{keys} without affecting output in ideal scenarios, followed by an exploration of practical approximations. To preserve the order of tokens, we focus on merging adjacent \textit{values}.

There is an ideal way to substitute two adjacent \textit{values} $\mathbf{v}_1,\mathbf{v}_2$ with $\mathbf{\tilde v}$ in $\mathbf{V}_t= [\mathbf{v}_1,\mathbf{v}_2\cdots,\mathbf{v}_t]$ without changing the output when we want to compress $\mathbf{k}_1,\mathbf{k}_2$ in \textit{key} cache $\mathbf{K}_t=[\mathbf{k}_1,\mathbf{k}_2,\cdots,\mathbf{k}_t]$ by keeping $\mathbf{k}_2$ and discarding $\mathbf{k}_1$. 
For easy of presentation, we assume the initial two positions are to be compressed.
To draw the formulation of $\mathbf{\tilde v}$, we consider the equation of outputs before and after the compression
\begin{align*}
&\mathrm{softmax}
\left(\mathbf{q}_t^\top [\mathbf{k}_1,\mathbf{k}_2,\mathbf{k}_3,\cdots,\mathbf{k}_t ]/\sqrt d\right)^\top
[\mathbf{v}_1,\mathbf{v}_2,\mathbf{v}_3,\cdots,\mathbf{v}_t]
\\
&=
\mathrm{softmax}
\left(\mathbf{q}_t^\top[\mathbf{k}_2,\mathbf{k}_3,\cdots,\mathbf{k}_t ]/\sqrt d\right)^\top
[\mathbf{\tilde v},\mathbf{v}_3,\cdots,\mathbf{v}_t].
\end{align*}
The equation gives
\begin{align*}
\mathbf{\tilde v} =&
\left(
1-\frac{e^{\mathbf{q}_t^\top \mathbf{k}_1/\sqrt d}}
{\sum_{i=1}^t e^{\mathbf{q}_t^\top \mathbf{k}_i/\sqrt d}}
\right)
\left(
e^{\mathbf{q}_t^\top (\mathbf{k}_1-\mathbf{k}_2)/\sqrt d}\mathbf{v}_1+\mathbf{v}_2
\right) \\
&- \frac{e^{\mathbf{q}_t^\top (\mathbf{k}_1-\mathbf{k}_2)/\sqrt d}}
{\sum_{i=1}^t e^{\mathbf{q}_t^\top \mathbf{k}_i/\sqrt d}}
\sum_{i=3}^t e^{\mathbf{q}_t^\top \mathbf{k}_i/\sqrt d}\mathbf{v}_i.
\end{align*}

We approximate the formula due to its $O(td)$ time complexity, omitting the terms with $\mathbf{v}_3, \ldots, \mathbf{v}_t$ and normalizing to prevent information decay. This yields the approximation
\[\mathbf{\tilde v}
\approx \frac{e^{\mathbf{q}_t^\top \mathbf{k}_1/\sqrt d}  }{
e^{\mathbf{q}_t^\top \mathbf{k}_1/\sqrt d} +
e^{\mathbf{q}_t^\top \mathbf{k}_2/\sqrt d} 
} \mathbf{v}_1 +
\frac{e^{\mathbf{q}_t^\top \mathbf{k}_2/\sqrt d}  }{
e^{\mathbf{q}_t^\top \mathbf{k}_1/\sqrt d} +
e^{\mathbf{q}_t^\top \mathbf{k}_2/\sqrt d} 
} \mathbf{v}_2,\]
which is a convex combination of $\mathbf{v}_1, \mathbf{v}_2$.
To minimize the approximation error, the attention weights of the evicted token might be kept as low as possible.
Motivated by this approximation, we propose WeightedKV to compress the KV cache by consolidating \textit{values} and evicting \textit{keys}.

\begin{table*}[t]
\small
\caption{Perplexity on PG19, OpenWebText2, ProofPile and ArXiv}
\centering
\resizebox{0.8\textwidth}{!}{
\begin{tabular}
{@{}l@{\hspace{0.5cm}}@{\hspace{0.5cm}}c@{\hspace{0.5cm}}c@{\hspace{0.5cm}}c@{\hspace{1cm}}c@{\hspace{0.5cm}}c@{\hspace{0.5cm}}c@{\hspace{1cm}}c@{\hspace{0.5cm}}c@{\hspace{0.5cm}}c@{\hspace{1cm}}c@{\hspace{0.5cm}}c@{\hspace{0.5cm}}c@{}}
\toprule
& \multicolumn{3}{c}{\hspace{-3.2em}\textbf{PG19}} & \multicolumn{3}{c}{\hspace{-3.4em}\textbf{OpenWebText2}} & \multicolumn{3}{c}{\hspace{-3.5em}\textbf{ProofPile}} & \multicolumn{3}{c}{\textbf{ArXiv}}\\ 
\cmidrule(r{28pt}){2-4} 
\cmidrule(r{28pt}){5-7}
\cmidrule(r{28pt}){8-10}
\cmidrule{11-13}
Context & 4k & 8k & 16k & 4k & 8k & 16k & 4k & 8k & 16k & 4k & 8k & 16k\\ 
\midrule
FullKV & 6.84  & - & -
& 5.44 & - & - 
& 2.51 & - & - 
& 3.04 & - & - \\ 
\arrayrulecolor{black}\midrule
\multicolumn{13}{c}{Cache Size = 1024} \\
\arrayrulecolor{black!20}\midrule
StreamingLLM & 7.19 &7.18  &7.20  
& 5.78 & 5.87 &  5.33
& 2.84 & 2.84 & 2.85  
& 3.48 & 3.45 & 3.49\\ 
TOVA & 7.00 & 7.05 & 7.14  
& 5.62 &5.73  &  5.24
&  2.64 & 2.67 & \textbf{2.72}  
& 3.22 & 3.24 & \textbf{3.30}\\ 
H2O & 7.06 & 7.06 & 7.24  
& \textbf{5.60} & 5.71 &  5.24
& \textbf{2.63} & {2.66} & 2.76 
& \textbf{3.20} &3.19  & 3.34\\ 
CaM &7.19&7.18&7.19
&5.85&5.89&5.46
&2.83&2.84&2.85
&3.48&3.48&3.49\\
WeightedKV &\textbf{6.98}  &\textbf{6.99}  & \textbf{7.10}
&\textbf{5.60}   & \textbf{5.68} &  \textbf{5.21}
& \textbf{2.63} & \textbf{2.64} & \textbf{2.72}
& \textbf{3.20} & \textbf{3.17} & \textbf{3.30}\\ 
\arrayrulecolor{black}\midrule
\multicolumn{13}{c}{Cache Size = 256} \\
\arrayrulecolor{black!20}\midrule
StreamingLLM & 7.99 & 8.00 &  7.99
& 6.67 & 6.66 & 6.05 
& 3.61 & 3.61 & 3.59 
& 4.51 &4.47  & 4.52\\ 
TOVA & 7.70 & 7.83 & 7.97 
& 6.36  &6.45  &  6.11
& 3.30 & 3.38 & 3.48 
& 4.12 &4.29 & 4.69\\ 
H2O &7.85  & 8.14 & 8.44 
& 6.50  & 6.64 &  6.23
& 3.36 & 3.60 & 3.83
& 4.13 & 4.38 & 4.77\\ 
CaM &7.98&7.97&7.99
& 6.70& 6.78 & 6.07
&3.62&3.63&3.61
&4.53& 4.53& 4.54\\
WeightedKV & \textbf{7.49} & \textbf{7.61} & \textbf{7.87} 
&\textbf{6.27}  &\textbf{6.37}  &  \textbf{5.97}
& \textbf{3.13} & \textbf{3.23} & \textbf{3.39} 
& \textbf{3.86} & \textbf{3.98} & \textbf{4.30} \\ 
\arrayrulecolor{black}\bottomrule
\end{tabular}
}
\label{tab: pp}
\end{table*}

\section{Methodology}
\label{sec: Methodology}

This section demonstrates the use of WeightedKV for compressing the KV cache. We first explain the selection and merging of \textit{values} for a single step, followed by a detailed illustration of the complete algorithm.

\begin{figure}
\centering
    \includegraphics[height=4.5cm]{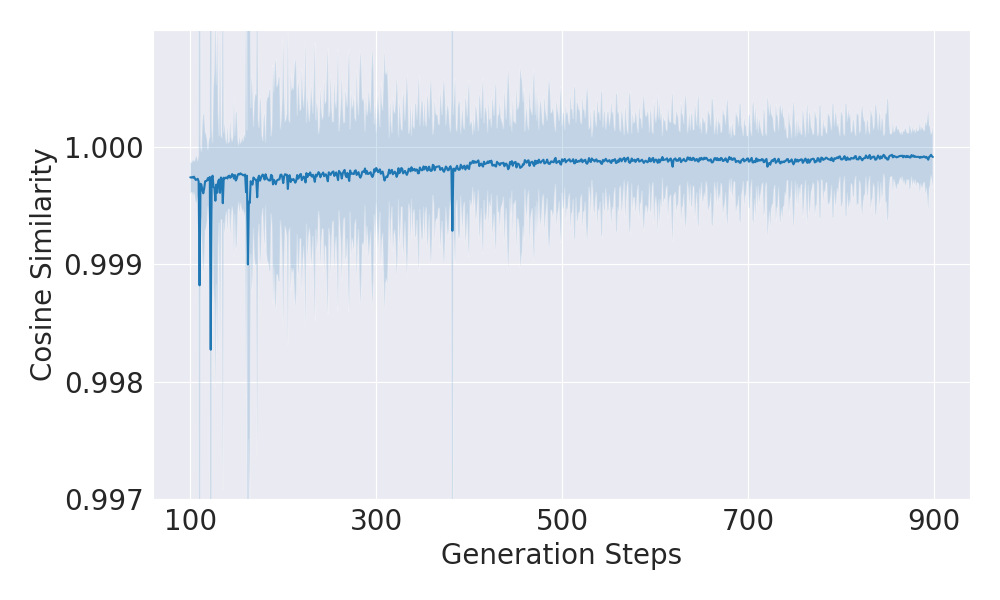}
    \vspace{-3mm}
    \caption{Cosine similarity between attention weights with merging \textit{values} and without merging \textit{values} at step 100 on the books from PG19.}
    \label{fig:cos_sim}
\end{figure}

In a single-step merge, we use average attention scores to weigh and consolidate \textit{values}. 
Following the analysis in section \ref{sec: ideal merge}, we first merge the \textit{value} with the least weight into its neighboring \textit{value}. 
To investigate the perturbation caused by merging, we merge the token with the lowest average attention with its adjacent right token for PG-19 test set samples at generation step 100, then calculate the attention scores for the next 800 steps. After that, we compute the attention scores on full attention and determine the cosine similarity to the previous scores.
Fig. \ref{fig:cos_sim} shows the cosine similarity between the attention scores over the next 800 steps, with standard variance represented as error bars.
The average cosine similarity is extremely close to 1, indicating that merging \textit{values} by such a convex combination has a minimal impact on the attention score of other tokens in the future. 
This experimental validation shows that our merging way has a negligible effect on subsequent inference steps.

\begin{figure}
\centering
    \includegraphics[height=5.3cm]{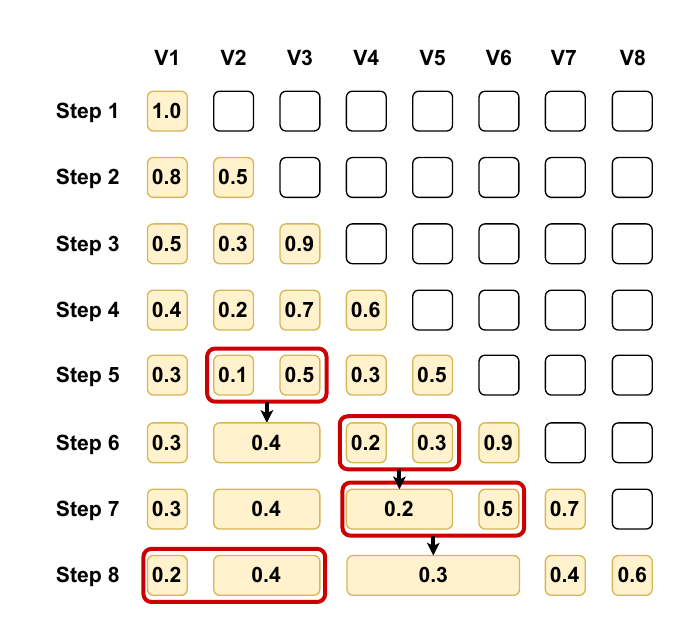}

    \vspace{-2mm}
    \caption{Compression process on a toy attention map with a maximum cache size of 4. Numbers in blocks represent average attention scores of tokens, while the red boxes indicate the \textit{values} to be merged.}
    \label{fig:merge}
\end{figure}

\begin{algorithm}[t]
\caption{WeightedKV}
\label{alg:algorithm}
\textbf{Input}: $\mathbf{x}_1,\cdots,\mathbf{x}_t$, $m$ 
\begin{algorithmic}[1] 
\STATE $\mathbf{a},\mathbf{n}$ are empty lists.
\FOR{$i$ from $1$ to $t$}
\STATE Get $\mathbf{k}_t,\mathbf{q}_t,\mathbf{v}_t,\mathbf{K}_t,\mathbf{V}_t,\mathbf{A}_t$ from $\mathbf{x}_t$. 
\STATE 
$\mathbf{a}=\left[\mathbf{a},0\right]+\mathbf{A}_t.$
\STATE 
$\mathbf{n}=\left[\mathbf{n},0\right]+\mathbf{1}.$
\IF{Length of $\mathbf{V}_t$ exceeds $m$}
\STATE $\overline{\mathbf{a}}=\mathbf{a}/\mathbf{n}$.
\STATE $j =\mathop{\arg\min}_{j\in\{0,\cdots, m-1\}} \overline{\mathbf{a}}[j]$ 
\STATE
Replace $\mathbf{V}_t[j:j+2]$ with 
$\frac{\overline{\mathbf{a}}[j]\mathbf{V}_t[j]+\overline{\mathbf{a}}[j+1]\mathbf{V}_t[j+1]}
{\overline{\mathbf{a}}[j]+\overline{\mathbf{a}}[j+1]}$.
\STATE 
Remove $\mathbf{K}_t[j], \mathbf{a}[j], \mathbf{n}[j]$.
\ENDIF
\ENDFOR
\end{algorithmic}
\end{algorithm}

The complete compression procedure of WeightedKV is outlined in Algorithm \ref{alg:algorithm}.
Our algorithm dynamically updates the accumulated attention weights and the calculation times for each key-value pair in lists $\mathbf{a},\mathbf{n}$ respectively during generation. 
When the cache reaches its maximum capacity $m$, historical average attention scores are calculated as $\mathbf{a}/\mathbf{n}$, which are weights for merging \textit{values} and evicting \textit{keys}. 
The \textit{value} corresponding to the smallest weight is selected to merge into its right adjacent \textit{value}.
The algorithm will merge two \textit{values} in the \textit{value} cache by convexly combining them weighted on historical average attention scores and remove the \textit{key} and item corresponding smaller weights from both the \textit{key} cache and the storage list. 
Fig. \ref{fig:merge} depicts a compression process by our algorithm on a toy average attention map with a maximum cache size of 4.
Numbers in blocks indicate the average attention scores of tokens. 
At generation step 5, the cache exceeds its capacity, triggering compression. Since $\mathbf{v}_2$ has the lowest average attention score of 0.1, the algorithm will replace $\mathbf{v}_2,\mathbf{v}_3$ by $\mathbf{v}_2/6+5\mathbf{v}_3/6$ given the $\mathbf{v}_3$'s score 0.5 and remove $\mathbf{k}_2$ at the same time. By repeating this procedure, after the compression at step 8, the cache maintains $\mathbf{v}_3,\mathbf{v}_6,\mathbf{v}_7,\mathbf{v}_8$ where $\mathbf{v}_3,\mathbf{v}_6$ contains the information of 3 single \textit{values} respectively.

\section{Experiments}

\begin{figure}[t]
    \centering
    \includegraphics[height=5cm]{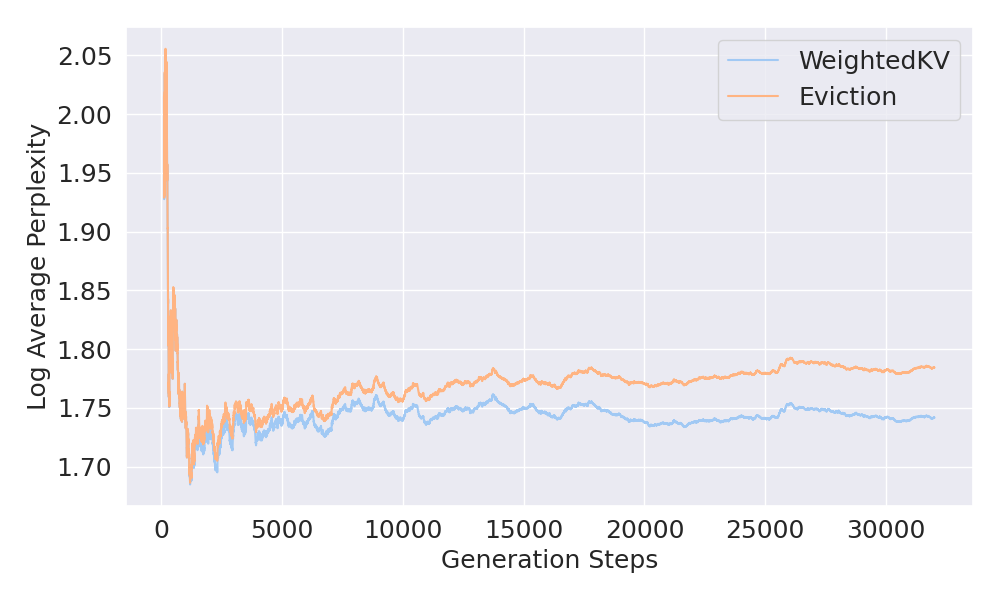}

    \vspace{-2mm}
    \caption{
    Comparison between WeightedKV and its eviction variant.}
    \label{fig:ablation}
\end{figure}

We assess WeightedKV's performance on four datasets that are commonly used for evaluating long text generation. Then we verify the efficacy of merging \textit{values} by an ablation study.

\subsection{Long Context Language Modeling}

We compare WeightedKV with five baseline methods. Three of them are efficient eviction-based methods: StreamingLLM \cite{xiao2023efficient}, H2O \cite{zhang2024h2o}, and TOVA \cite{oren2024transformers}. The remaining two are the efficient merging-based method CaM\cite{zhang2024cam} and the full attention method.
We report sliding window perplexity with window sizes ranging from 4k to 16k and strides of half of the windows on four datasets using Llama-2-7B\cite{touvron2023llama}. 
We assess perplexity on the complete test set of PG19\cite{rae2019compressive}, which contains 100 books with an average length of 70k. We randomly select 10 samples from the ProofPile\cite{proofpile2022} test set and ArXiv\cite{gao2020pile} train set, averaging 128k tokens for evaluation. For OpenWebText2\cite{gao2020pile}, we randomly select 100 samples averaging 18k tokens from its test set. We retain 4 initial tokens in cache and consider cache sizes of 1024 and 256, keeping 508 and 124 recent tokens for each size.

The results in Table \ref{tab: pp} indicate that, with a cache size of 1024, WeightedKV outperforms all other efficient methods on the PG19 dataset across all context lengths. 
On the other three datasets, our approach is tied for best with H2O or TOVA.
WeightedKV outperforms all other methods across all datasets and context lengths with a smaller cache size 256. These results demonstrate that WeightedKV is an effective method for KV cache compression, particularly evident at smaller cache sizes.

\subsection{Ablation Study}

To validate the pivotal role of merging, we consider a pure evicting method that compresses the KV cache in the same manner as WeightedKV in all other aspects, but instead of merging, it evicts \textit{values}.
This modification gives a targeted analysis of the impact of token merging on model performance.
In Fig. \ref{fig:ablation}, we label our method as ``WeightedKV" and the modified method as ``Eviction", showing their log average perplexity on the first sample from the PG19 test set truncated at 32k. The comparison between WeightedKV, which merges \textit{values} and evicts \textit{keys} (blue line), and the pure evicting method (orange line) demonstrates 
that our merging approach significantly improves perplexity, confirming that merging is essential to our method.

\section{Conclusion}
Our findings underline the importance of rethinking conventional cache eviction strategies. The WeightedKV method that discards unimportant token \textit{keys}, while convexly merging their \textit{values} into neighboring tokens, weighted by their average attention scores. Evaluations of our approach on various datasets demonstrate its superiority over existing methods.

\newpage
\bibliographystyle{IEEEtran}
\bibliography{main}


\end{document}